\documentclass[10pt,twocolumn,letterpaper]{article}
\usepackage{graphicx}
\usepackage{multirow}
\usepackage{amsmath}
\usepackage{booktabs}
\usepackage{makecell}

\usepackage{graphicx}  
\usepackage{array}     

\usepackage[pagenumbers]{cvpr} 

\definecolor{cvprblue}{rgb}{0.21,0.49,0.74}
\usepackage[pagebackref,breaklinks,colorlinks,allcolors=cvprblue]{hyperref}

\def\paperID{861} 
\def\confName{CVPR}
\def\confYear{2025}

\title{ST-Align: A Multimodal Foundation Model for Image-Gene Alignment in Spatial Transcriptomics}

\author{
Yuxiang Lin$^{1}$\thanks{These authors contributed equally to this work.}, 
Ling Luo$^{1}$\footnotemark[1], 
Ying Chen$^{2}$\footnotemark[1], 
Xushi Zhang$^{1}$,
Zihui Wang$^{2}$,
Wenxian Yang$^{4}$\thanks{Corresponding author.},\\
Mengsha Tong$^{1,3}\footnotemark[2]$,
Rongshan Yu$^{1,2}\footnotemark[2]$
\\
$^{1}$National Institute for Data Science in Health and Medicine, Xiamen University, Xiamen, China \\
$^{2}$School of Informatics, Xiamen University, Xiamen, China \\
$^{3}$School of Life Sciences, Xiamen University, Xiamen, China \\
$^{4}$Aginome Scientific, Xiamen, China  \\
{\tt\small linyuxiang@stu.xmu.edu.cn, luoling2001@stu.xmu.edu.cn, rsyu@xmu.edu.cn}
}

\begin{document}
\maketitle
\begin{abstract}
Spatial transcriptomics (ST) provides high-resolution pathological images and whole-transcriptomic expression profiles at individual spots across whole-slide scales.
This setting makes it an ideal data source to develop multimodal foundation models. 
Although recent studies attempted to fine-tune visual encoders with trainable gene encoders based on spot-level, the absence of a wider slide perspective and spatial intrinsic relationships limits their ability to capture ST-specific insights effectively. 
Here, we introduce ST-Align, the first foundation model designed for ST that deeply aligns image-gene pairs by incorporating spatial context, effectively bridging pathological imaging with genomic features. 
We design a novel pretraining framework with a three-target alignment strategy for ST-Align, enabling (1) multi-scale alignment across image-gene pairs, capturing both spot- and niche-level contexts for a comprehensive perspective, and 
(2) cross-level alignment of multimodal insights, connecting localized cellular characteristics and broader tissue architecture. 
Additionally, ST-Align employs specialized encoders tailored to distinct ST contexts, followed by an Attention-Based Fusion Network (ABFN) for enhanced multimodal fusion, effectively merging domain-shared knowledge with ST-specific insights from both pathological and genomic data.
We pre-trained ST-Align on 1.3 million spot-niche pairs and evaluated its performance through two downstream tasks across six datasets, demonstrating superior zero-shot and few-shot capabilities. 
ST-Align highlights the potential for reducing the cost of ST and providing valuable insights into the distinction of critical compositions within human tissue. 

\end{abstract}

\section{Introduction}
\label{sec:intro}

\begin{figure}[th]
  \centering
   \includegraphics[width=\linewidth]{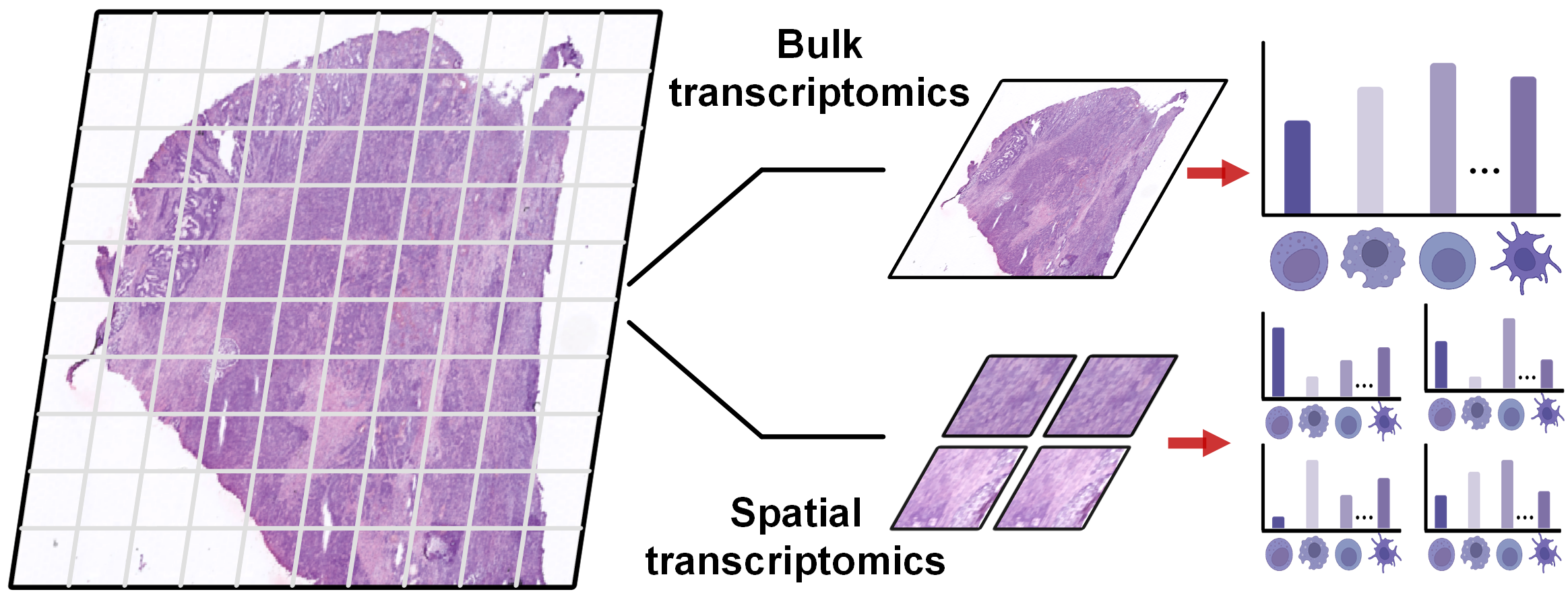}
   \caption{\textbf{Comparison between WSI-Bulk Transcriptomics and ST Data.} 
   ST enables the integration of high-resolution histopathological images with whole-transcriptomic gene expression profiles at the level of individual spots across the entire slide. 
   In contrast, bulk transcriptomics averages gene expression across heterogeneous cell populations, lacking spatial resolution and the ability to correlate gene expression with specific regions or patches within WSIs.}
   \label{fig_1_ST}
\end{figure}

In modern healthcare, exploring the homogeneous or heterogeneous cellular components within spatial niches is critical \cite{tong2023prioritizing,de2023evolving,bejarano2021therapeutic,chen2023general,xiong2024mome,jaume2024multistain}. Traditionally, hematoxylin and Eosin (H\&E) stained-whole slide images (WSIs) and bulk gene expression profiles (GEPs) have been widely employed to investigate the cellular morphology and intrinsic genetic statuses of tissues \cite{niazi2019digital,lu2024visual,chen2024uni,tian2024prediction,xu2024gigapath,wang2024pathology,song2024morphological,zhang2024pathology,jaume2024modeling}. 
However, bulk GEPs do not provide sufficient genetic context corresponding to the high resolution of WSIs, hindering researchers from exploring the characteristics of niches with distinct genetic profiles\cite{jaume2024transcriptomics,ding2023pathology,zhangprototypical,tang2024feature,chen2024survmamba}. 

ST is a novel technology that combines high-resolution imaging with high-throughput sequencing \cite{chen2015spatially,wang2018three}. 
In ST, thousands of spots, each with a radius of 55 \(\mu m\), are placed on a chip measuring 6.5 \(mm\) \(\times\) 6.5  \(mm\). 
This design facilitates the capture of corresponding H\&E images and GEPs within a spatial context, ensuring fine-grained alignment between histological morphology and molecular features across numerous sub-tile regions (shown in Figure~\ref{fig_1_ST}), which highlights ST an ideal source for paired pathological images and genes.

Recent research efforts have focused on collecting these novel and valuable ST to advance this field. 
In addition, inspired by the success of vision-language models \cite{christensen2024vision,sun2024pathasst}, researchers fine-tuned the original CLIP framework with spots from ST and explored the construction of image-gene multimodal models.
However, modeling ST with CLIP or PLIP immediately poses challenges including (1) overlooking the inherent spatial relationships between spots and corresponding broader niches, leading to limited modeling of ST and loss of valuable insights; (2) pre-trained visual encoders struggle to adapt ST images of varying scales, while gene encoders trained from scratch may exhibit limited generalizability.



In this study, we design a pretraining paradigm and propose the first image-gene foundation model named \textbf{ST-Align} for ST, aligning pathology image-gene relationships across multiple spatial scales and broadening the context of ST modeling. 
(1) We focus on spot and niche simultaneously, employing a three-target alignment strategy to achieve comprehensive image-gene alignment and broader perceive of structural characteristics within ST. 
Specifically, the alignment objectives span three components: image-gene alignment at the spot level, image-gene alignment at the niche level, and a further alignment of the integrated multimodal features from spots and niches. 
(2) We design specialized encoders for distinct context in ST, followed with
an Attention-Based Fusion Network (ABFN) to fuse visual and genetic feature. 
This approach not only enhances adaptability to images and genes of varying sizes, but also incorporates domain-common knowledge from previous well-established pre-trained models alongside ST-specific insights from additional encoders. 

To develop ST-Align, we curated 1.3 million image-gene pairs, each with corresponding spot-level and niche-level information, to pre-train the model and evaluated its performance on two downstream tasks: spatial cluster identification and gene prediction across six in-the-wild datasets. 
To summarize, our contributions are: 
(1) we introduced a novel pre-training paradigm with a three-target alignment strategy and trained ST-Align on 1.3 million image-gene pairs. To the best of our knowledge, ST-Align is the first image-gene foundation model for ST, broadening the scope of ST applications.
(2) We designed specialized encoders to capture distinct contextual features in ST, followed by an ABFN module to fuse multimodal data, integrating domain-shared knowledge with ST-specific insights from both visual and genetic features.
(3) A series of downstream experiments, including niche-level spatial clustering and spot-level gene expression prediction, conducted on six benchmark datasets, show the generalizability of ST-Align.
\section{Related Work}
\label{sec:Related Work}



\subsection{Multimodal Foundation Model}

Multiple pathological image-text pair datasets have emerged as foundational resources for constructing multimodal foundation models in medical ares.
The OpenPath dataset provides a comprehensive resource, featuring 116,504 image-text pairs from Twitter posts across 32 pathology subspecialties, facilitating the fine-tuning of a PLIP foundation model to enhance diagnosis, knowledge sharing, and pathological education \cite{huang2023visual,schuhmann2022laion,yin2024prompting,Li_2024_CVPR,sun2025pathmmu}.
Quilt-1M serves as another significant source, yielding over 1 million paired samples that have been utilized for fine-tuning a pre-trained CLIP model, demonstrating its performance across diverse sub-pathologies and cross-modal retrieval tasks \cite{ikezogwo2024quilt}. 
A recent visual-language foundation model, CONCH, was developed using various pathological images and biomedical text, incorporating over 1.17 million image-caption pairs through task-agnostic pretraining, achieving state-of-the-art (SOTA) performance across 14 diverse benchmark tasks \cite{lu2024visual}.
PathAsst and PathCLIP were trained on over 207K high-quality pathology image-text pairs from public sources, facilitating advancements in the interpretation of pathology images, as well as in diagnosis and treatment processes \cite{sun2024pathasst}.
Collectively, these multimodal datasets provide external insights into understanding and uncovering the information contained in pathological images, thereby facilitating improvements in performance across various downstream tasks, including diagnosis and clinical report synthesis.

\subsection{Foundation Models for WSI and GEP}

\textbf{Pathological Foundation Model: }
The recent advances in the area of foundation model of WSI had gain significant traction in pathology.
The previous pathological foundation model combined with self-supervised learning and swin Tranformer and it was trained on TCGA dataset, which contain more than 10 thousand of WSIs \cite{wang2022transformer}. 
Existing SOTA methods was developed on exceeding 1 million WSIs from diverse sources and with rich biomedical text and other modality and this novel adopt the new contrastive leanring strategy and efficient attention mechanism, which archive inspired performance in more than 15 diverse downstream stream tasks \cite{chen2024uni,tian2024prediction,wang2024pathology}.

\noindent \textbf{Genetic Foundation Model:}
In the area of transcriptome, existing foundation approaches focus on the single-cell transcriptomics data and apply the reconstruction loss to guide the model learning the intrinsic gene expression pattern \cite{yang2022scbert,cui2023scGPT,hao2024large}. 
It can also be pushed one step further to involve other modality in extending the biological insights \cite{zhao2023large,bian2024scmulan,2024wencellplm,khwaja2024celle}.
Collectively, these model demonstrate impressive performance in solving multimodal downstream tasks, as well as in bring novel intrinsic biological insights.

\subsection{Image-Gene Paired Datasets}
Previous image-gene datasets were based on pairwise WSI and bulk transcriptomic GEPs derived from the same patient.
Specifically, the bulk GEP was a vector containing 19,000 protein-coding genes for individual patient samples, corresponding to a gigapixel WSI. 
The rise of ST has spurred the development of various datasets focused on fine-grained transcriptomic analysis in tissue. 
ST allows researchers to obtain paired pathological images and transcriptome at a single spot, each with a 55 µm diameter, with thousands of spots arranged across tissue slices. 
Recent databases include CROST \cite{wang2024crost}, SODB \cite{yuan2023sodb}, STOmicsDB \cite{xu2024stomicsdb}, Aquila \cite{zheng2023aquila}, and the Museum of Spatial Transcriptomics \cite{moses2022museum}. 
These databases primarily focus on collecting normal, disease and cancerous ST data, providing valuable insights into the spatial distribution of gene expression in tissue samples.
In additional, HEST-1k \cite{jaume2024hest} and STimage-1K4M \cite{chen2024stimage1k4m}, offer paired image and gene expression data, making them especially ideal source for bridging the gap between visual information and genetic expression in pathological area.

\subsection{Downstream Tasks in ST}

\textbf{Representation Learning and Clustering.}
Learning informative representation is a important task in ST.
This process involves compact WSI and GEP, capturing the intrinsic features of the underlying biological processes. 
The results can be applied in distinguishing spatial clusters, where tissue regions are grouped based on shared characteristics captured in the embeddings \cite{hu2021spagcn,hu2024unsupervised,ma2024accurate}. 
Clustering is a basic task and allow researchers to explore tissue heterogeneity and identify distinct spatial niches that represent the different cellular functions or disease states.

\noindent \textbf{Gene Expression Enhancement and Prediction.}
Another key task in ST is learning the relationship between pathological images and gene expression, enabling the prediction of gene expression directly from the images. 
This approach has the potential to reduce the need for costly and time-consuming library preparation and sequencing \cite{zhang2024inferring}.
Additionally, improving the quality of sequencing and increasing the resolution of GEP through high-resolution imaging techniques offers a more detailed understanding of spatial patterns within tissue samples \cite{wang2022sprod,si2024ficture,benjamin2024multiscale}. 
It leading to improved accuracy in analyzing gene expression spatial distributions among heterogeneous spatial niches.
\begin{figure*}[t]
  \centering
   \includegraphics[width=18cm,height=15cm]{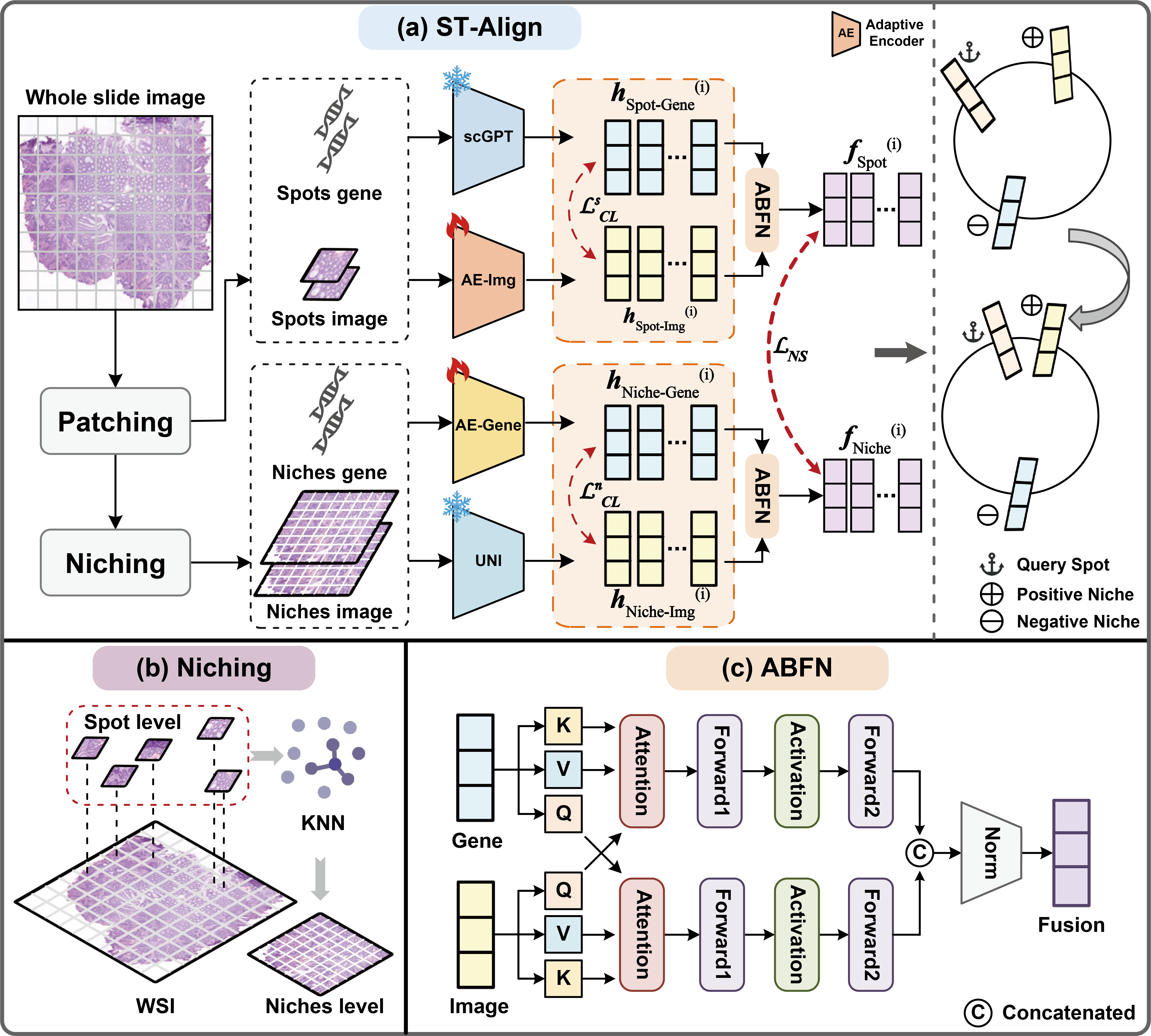}
   \caption{\textbf{Overview of ST-Align Architecture.} (a) Paired WSI and GEP data are segmented into spot-level patches, which are then grouped into niche-level data. A compressed feature for each paired spot-level gene and niche-level image is encoded using a feature extractor pretrained on a large dataset, while spot-level images and niche-level genes are encoded using trainable encoder. In addition, We not only aligned image feature and gene feature at spot-level and niche-level, but also aligend spot-niche fusion feature. (b) The KNN algorithm is used to cluster spot-level data to obtain niche-level data. (c) Attention based fusion network.}
   \label{fig:method-one}
\end{figure*}

\section{Methods}
\label{sec:methods}
Here, we present ST-Align, the first image-gene foundation model with a novel pre-training paradigm specifically designed for ST.  The model architecture is illustrated in Figure~\ref{fig:method-one}.  First, we represent ST as a multi-level spatial structure in Section~\ref{sec:ST Structure}. Next, we detail the specialized image and gene encoders for ST in Section~\ref{sec:ST Encoder}. Then, in Section~\ref{sec: ABFN}, we present the Attention-Based Fusion Network (ABFN) for integrating visual and genetic features. Finally, the alignment objectives for ST-Align pretraining are introduced in Section~\ref{sec:Alignment Objectives}.

\subsection{Muti-level Spatial Structure of ST}
\label{sec:ST Structure}

Recognizing the spatial heterogeneity of ST, we represent it as a muti-level spatial structure with spot-level and niche-level. Spots reflect microscopic information in a small region, while niches represents a larger functional area composed of multiple adjacent spots. Given a histology slide \( X_i \in \mathbb{R}^{d_x \times d_y \times 3} \), we not only tessellate it into \textbf{spot-level} patches based on the coordinates of the spatial transcriptome sequencing points \( S_i = \{ \mathbf{s}_i^1, \ldots, \mathbf{s}_i^{N_i} \} \) with \(  \mathbf{s}_i^n  \in \mathbb{R}^{W_s \times H_s \times 3} \), but also according to the KNN algorithm (\cref{sec:Related Work}) based on Euclidean distance (\cref{eq:distance}), the sequencing points are clustered to segment the \textbf{niche-level} patches \( G_i = \{ \mathbf{g}_i^1, \ldots, \mathbf{g}_i^{N_i} \} \) with \(  \mathbf{g}_i^n  \in \mathbb{R}^{W_g \times H_g \times 3} \).

\begin{equation}
  L_2(x_i, x_j) = \left( \sum_{l=1}^{n} \left| x_i^{(l)} - x_j^{(l)} \right|^2 \right)^{\frac{1}{2}},
  \label{eq:distance}
\end{equation}

\noindent where \(x_i, x_j\) represent two sequencing points; \(n = 2\) represents a two-dimensional space, and \(x_i\) and \(x_j\) denote the coordinate values of \(x_i^{(l)}\) and \(x_j^{(l)}\) in the LTH dimension, respectively.

\noindent Given a set of gene expression from spatial transcriptome sequencing \( G_i \in \mathbb{R}^{N_g} \) which results corresponding to histology slide \( X_i \). Spot-level gene expression values \( Q_i = \{ \mathbf{q}_i^1, \ldots, \mathbf{q}_i^{N_i} \} \), where \( \mathbf{q}_i^n \in \mathbb{R}^{N_g} \), can be obtained for each sequencing point. For niche-level gene expression, we calculate the mean of the gene expression values (\cref{eq:get-niches-gene}) across all sequencing points within the niche-level cluster, it can be defined as \( P_i = \{ \mathbf{p}_i^1, \ldots, \mathbf{p}_i^{N_i} \} \), where \( \mathbf{p}_i^n \in \mathbb{R}^{N_g} \).

\begin{equation}
    p_i^n = \frac{1}{|S|} \sum_{j \in S} q_i^j,
    \label{eq:get-niches-gene}
\end{equation}

\noindent where n represents the index of the sequencing points and S represents the set of points within the niche-level cluster.

\subsection{ST Encoder}
\label{sec:ST Encoder}
\noindent \textbf{Image Encoding: } 
It is important to highlight that ST spot-level images are relatively small, measuring only 28×28 pixels, which presents a challenge for traditional visual foundation models to effectively extract meaningful information. To address this, we employ a custom-designed adaptive encoder to extract features from these diminutive spot images. Here, we select ResNet-50~\cite{He2015} as the encoder, referred to as AE-Img, and utilize a from-scratch training approach. AE-img is tasked with capturing the fine-grained features of a given set of spot-level patches \( S_i \), with the output defined as follows:

\begin{equation}
   X_s = \{ R_1, R_2, \ldots, R_{N_i} \} = \text{AE-Img}(S_i; \theta_t^{\text{resnet}}),
  \label{eq:res_emb}
\end{equation}

\noindent where \(R_{n} \in \mathbb{R}^d \) is the vector after embedding and \(\theta_t^{\text{resnet}}\) denotes the parameters of the AE-Img Encoder.

For niche-level images, we preprocess them to a resolution of 224×224 pixels and then employ the pretrained pathology image encoder. Given a sequence of niche-level images \( G_i\), the output of the UNI model can be defined as:
\begin{equation}
   X_g = \{ E_1, E_2, \ldots, E_{N_i} \} = \text{UNI}(G_i; \theta_t^{\text{uni}}),
  \label{eq:uni_emb}
\end{equation}

\noindent where \(\|\mathbf{X_g}\| = \|\mathbf{X_s}\|\), for the same WSI, \(E_i\) and \(R_i\) correspond one-to-one, and \(E_{n} \in \mathbb{R}^d \) is the embedding.

\noindent \textbf{Gene Encoding:}
Existing genetic foundation models are typically trained on single-cell transcriptome data. 
However, the genetic data for individual spots generally represents 2 to 10 cells in ST, resulting in further divergence from single-cell genetic data distributions.

To address this, we utilize a pretrained model to capture information at the spot level, while designing an adaptive encoder to model gene expression at the niche level, referred to as AE-Gene.
In addition, scGPT~\cite{cui2023scGPT}, a generative pre-trained transformer trained on a repository of over 33 million cells, was leveraged to extract features from a given set of spot-level genes \( Q_i \), as shown in \cref{eq:scgpt}.
\begin{equation}
   G_s = \{ S_1, S_2, \ldots, S_{N_i} \} = \text{scGPT}(Q_i; \theta_t^{\text{scgpt}}),
  \label{eq:scgpt}
\end{equation}
\noindent where \( \theta_t^{\text{scgpt}} \) represents the pretrained parameters of the scGPT model, and \( S_{n} \in \mathbb{R}^d \) is the spot-level gene embedding.

As for AE-Gene, we select Transformer Encoder~\cite{vaswani2017attention} serves as a trainable module to learn niche-level gene features.
\begin{equation}
   G_g = \{ T_1, T_2, \ldots, T_{N_i} \} = \text{AE-Gene}(P_i; \theta_t^{\text{trans}}),
  \label{eq:scgpt}
\end{equation}
\noindent where \( T_{n} \in \mathbb{R}^d \) is the embedded vector, and \( \theta_t^{\text{trans}} \) represents the parameters of Transformer.

\subsection{Attention-Based Fusion Network}
\label{sec: ABFN}
After extracting features using the image and gene encoder, we employ a cross-attention mechanism to facilitate interaction between image features \( F^I = \{ \mathbf{fI}_1, \ldots, \mathbf{fI}_{N_i} \} \) with \(F^I \in \mathbb{R}^{d}\) and gene features \( F^G = \{ \mathbf{fG}_1, \ldots, \mathbf{fG}_{N_i} \} \)  with \(F^G \in \mathbb{R}^{d}\), thereby enhancing image features with gene context, the formulation of this interaction is as follows:

\label{sec:Multimodal Fusion}

\begin{equation}
    Z_i^{I} = \frac{\exp \left( \frac{( \mathbf{fI^\prime}_i \cdot W_q) \cdot (\mathbf{fG^\prime}_i \cdot W_k)^T}{\sqrt{d}} \right) \cdot \mathbf{fG^\prime}_i \cdot W_v}{\sum_j \exp \left( \frac{( \mathbf{fI^\prime}_j \cdot W_q) \cdot (\mathbf{fG^\prime}_j \cdot W_k)^T}{\sqrt{d}} \right)},
\end{equation}

\noindent where \(\mathbf{fI_i} \in \mathbb{R}^{d} \rightarrow \mathbf{fI_i^\prime} \in \mathbb{R}^{ s \times l}\), and \( W_q, W_k, W_v \in \mathbb{R}^{l \times l} \) are learnable embedding matrices.

Similarly, we enhance gene features by incorporating image context:
\begin{equation}
    Z_i^{G} = \frac{\exp \left( \frac{( \mathbf{fG^\prime}_i \cdot W_q) \cdot (\mathbf{fI^\prime}_i \cdot W_k)^T}{\sqrt{d}} \right) \cdot \mathbf{fI^\prime}_i \cdot W_v}{\sum_j \exp \left( \frac{( \mathbf{fG^\prime}_j \cdot W_q) \cdot (\mathbf{fI^\prime}_j \cdot W_k)^T}{\sqrt{d}} \right)},
\end{equation}

\noindent where \(\mathbf{fG_i} \in \mathbb{R}^{d} \rightarrow \mathbf{fG_i^\prime} \in \mathbb{R}^{ s \times l}\), and \( W_q, W_k, W_v \in \mathbb{R}^{l \times l} \) are learnable embedding matrices.

\noindent Finally, we merge the two enhanced feature vectors to obtain the multimodal representation. The formulation of this interaction is as follows:

\begin{equation}
    F_i = [ Z_i^{I} W_{I}; Z_i^{G} W_{G}],
\end{equation}

\noindent where \(W_{I}, W_{G} \in \mathbb{R}^{l \times \frac{l}{2}}\), \([\mathbf{\cdot}; \mathbf{\cdot}]\) denotes the concatenation operation, and \(F_i \in \mathbb{R}^{s \times l} \rightarrow F_i \in \mathbb{R}^{d}\).

\subsection{Alignment Objectives}
\label{sec:Alignment Objectives}

\noindent \textbf{Muti-level Image-Gene Alignment:} 
We align the embedding spaces of the slide and expression encoders through a symmetric cross-modal contrastive learning objective.  For a spot-level image embedding \(R_i\), given a set \( T = \{ a_1, \ldots, a_u \} \), where \(  \mathbf{T}  \in \mathbb{R}^{d} \) is a subset of spot-level gene expression embeddings containing one positive sample and \( u-1\) samples, we optimize:

\begin{align}
    \mathcal{L}_{CL}^s = & -\frac{1}{2} \log \left( \frac{\exp \left( \frac{R_i \cdot a_i^+}{\tau} \right)}{\sum_{j=1}^{u-1} \exp \left( \frac{R_i \cdot a_j^-}{\tau} \right)} \right) \nonumber \\
    & -\frac{1}{2} \log \left( \frac{\exp \left( \frac{a_i \cdot R_i^+}{\tau} \right)}{\sum_{j=1}^{u-1} \exp \left( \frac{a_i \cdot R_j^-}{\tau} \right)} \right),
    \label{eq:infonce}
\end{align}

\noindent where \(R_i\) and \(a_i\) are used as the query sample, \(a_i^+\) and \(R_i^+\) are the positive sample corresponding to the query, \(a_i^-\) and \(R_i^-\) are the negative sample, and \(\tau\) is the temperature coefficient used to regulate the distribution of similarity scores.

For niche-level image and gene expression embeddings, we adopt the same optimization objective to align them, denoted by the objective function \( \mathcal{L}_{CL}^{n} \).

\noindent \textbf{Spot-Niche Alignment:} Beyond traditional inter-modality alignment, we introduce an approach to align spot-level and niche-level feature embeddings, effectively increasing the receptive field at the spot level and enhancing the ability to capture the structural features of pathological images. 

\begin{equation}
    \mathcal{L}_{NS} = - \log \left( \frac{\exp \left( \frac{F_i^S \cdot F_i^{N+}}{\tau} \right)}{\sum_{j=1}^{u-1} \exp \left( \frac{F_i^S \cdot F_j^{N-}}{\tau} \right)} \right),
    \label{eq:infonce}
\end{equation}

\noindent where \(F_i^S, F_i^N \in \mathbb{R}^{d}\) are the feature embeddings of spot-level and niche-level, respectively, after multimodal fusion.

We optimize the above objectives with total loss \(\mathcal{L}\):

\begin{equation}
    \mathcal{L} = \lambda_{1} \mathcal{L}_{CL}^{s} + \lambda_{2} \mathcal{L}_{CL}^{n} + (1 - \lambda_{1} - \lambda_{2}) \mathcal{L}_{NS}.
    \label{eq:total-loss}
\end{equation}

Here, \( \lambda_{1} \) and  \( \lambda_{2} \) are hyperparameters that balances the contribution of each loss type.
\section{Experiments and results}

\begin{table*}[ht]
\centering
\small
\caption{\textbf{Performance of different foundation model embeddings in spatial clustering identification.} \textbf{G.} and \textbf{P.} refer to graph-based modality (transcriptomics) and path-based modality (WSIs), respectively. Best performance in \textbf{bold}, second best \underline{underlined}. The standard deviation is reported over five runs and evaluated using the ARI metric.
}
\label{tab:results-one}
\renewcommand{\arraystretch}{1.2}  
\setlength{\tabcolsep}{4pt}        
\begin{tabular}{l|cc|cccccc|c}
\hline
\multirow{2}{*}{\textbf{Model}} & \multicolumn{2}{c|}{\textbf{Modality}} & \multicolumn{6}{c|}{\textbf{Dataset}} & \multirow{2}{*}{\textbf{Overall}} \\

\cline{4-9}
 & \textbf{G.} & \textbf{P.} & 151507 & 151508 & 151509 & 151669 & 151670 & 151673 & \\
\hline

CTransPath & & $\checkmark$ & 0.0589\scriptsize± 0.030 & 0.0702\scriptsize± 0.036 & 0.0823\scriptsize± 0.034 & 0.0030\scriptsize± 0.007 & 0.0482\scriptsize± 0.006 & 0.2269\scriptsize± 0.013 & 0.0816 \\

UNI & & $\checkmark$ & 0.1056\scriptsize± 0.044 & 0.1068\scriptsize± 0.046 & 0.1647\scriptsize± 0.060 & 0.0022\scriptsize± 0.005 & 0.0642\scriptsize± 0.029 & 0.2101\scriptsize± 0.027 & 0.1089 \\

Prov-GigaPath & & $\checkmark$ & 0.1047\scriptsize± 0.021 & 0.0951\scriptsize± 0.030 & 0.1535\scriptsize± 0.067 & 0.0314\scriptsize± 0.039 & 0.0880\scriptsize± 0.006 & 0.1927\scriptsize± 0.018 & 0.1109 \\

Hibou & & $\checkmark$ & 0.0669\scriptsize± 0.033 & 0.0609\scriptsize± 0.034 & 0.0754\scriptsize± 0.040 & 0.0132\scriptsize± 0.029 & 0.0862\scriptsize± 0.003 & 0.2198\scriptsize± 0.010 & 0.0871 \\

CONCH & & $\checkmark$ & 0.1019\scriptsize± 0.022 & 0.1623\scriptsize± 0.039 & 0.1930\scriptsize± 0.064 & 0.0053\scriptsize± 0.009 & 0.0838\scriptsize± 0.005 & 0.2243\scriptsize± 0.024 & 0.1284 \\

\hline

Scanpy & $\checkmark$ & & 0.2184\scriptsize± 0.031 & 0.2246\scriptsize± 0.018 & 0.3902\scriptsize± 0.026 & \underline{0.2878}\scriptsize± 0.202 & 0.2334\scriptsize± 0.1635 & 0.2288\scriptsize± 0.027 & \underline{0.2639} \\

scFoundation & $\checkmark$ & & 0.2058\scriptsize± 0.021 & 0.2333\scriptsize± 0.021 & 0.3869\scriptsize± 0.027 & 0.2851\scriptsize± 0.061 & 0.2593\scriptsize± 0.060 & 0.1989\scriptsize± 0.031 & 0.2616 \\

scGPT & $\checkmark$ &  & 0.2483\scriptsize± 0.021 & 0.2592\scriptsize± 0.011 & 0.3282\scriptsize± 0.034 & 0.2115\scriptsize± 0.145 & \underline{0.2869}\scriptsize± 0.038 & \underline{0.2348}\scriptsize± 0.031 & 0.2615 \\

\hline

CLIP & $\checkmark$ & $\checkmark$ & \underline{0.2977}\scriptsize± 0.031 & \underline{0.3171}\scriptsize± 0.021 & 0.3747\scriptsize± 0.024 & 0.1136\scriptsize± 0.031 & 0.2277\scriptsize± 0.061 & 0.2058\scriptsize± 0.013 & 0.2561 \\

PLIP & $\checkmark$ & $\checkmark$ & 0.2707\scriptsize± 0.040 & 0.3010\scriptsize± 0.008 & \underline{0.4207}\scriptsize± 0.018 & 0.0918\scriptsize± 0.051 & 0.1790\scriptsize± 0.036 & 0.2267\scriptsize± 0.012 & 0.2483 \\

\textbf{ST-Align} \scriptsize & $\checkmark$ & $\checkmark$ & \textbf{0.3098}\scriptsize± 0.016 & \textbf{0.3319}\scriptsize± 0.035 & \textbf{0.4700}\scriptsize± 0.037 & \textbf{0.2956}\scriptsize± 0.100 & \textbf{0.3523}\scriptsize± 0.067 & \textbf{0.2783}\scriptsize± 0.014 & \textbf{0.3396} \\

\hline
\end{tabular}
\end{table*}

\label{sec:Experiments and results}


\subsection{Dataset and Implementation}

\noindent \textbf{Spot view data collection:} 
All image-gene pair data were derived from the public dataset STimage-1K4M~\cite{chen2024stimage1k4m}, which covers 11 tissue types and was sequenced using three distinct ST technologies. 
To ensure consistent scale in spot images, we retained only data from human tissues and sequenced with 10x Visium technology. 
Additionally, we filtered out WSIs with fewer than 50 spots, yielding a final dataset of 573 WSIs with 1.3 million spots.

\noindent \textbf{Niche view data collection:} 
For each individual spot in the dataset, we collected its correspond niche, defined as the three nearest neighboring spots that provide a larger-scale context. 
To approximate the niche-level transcriptomic GEP, we averaged the expression values of the three neighboring spots to simulate bulk transcriptomics at the niche level. 
Consequently, we constructed paired pathological and genetic data for each of the 1.3 million spots along with their corresponding niche.

\noindent \textbf{Implementation:} 
For AE-Gene, we use a 6-layer Transformer encoder with an 8-head attention mechanism, and set the dropout rate to 0.1.
During training, the learning rate was initialized at \(5 \times 10^{-4}\), with a cosine scheduler and linear warmup for gradual adjustment. 
We used the AdamW optimizer with a weight decay ranging from \(0.04\) to \(0.4\), following a cosine decay schedule.
The optimizer parameters include \(\epsilon = 1 \times 10^{-8}\) and \(\beta = (0.9, 0.999)\). 
Model training was distributed across 3 NVIDIA A800 GPUs, with synchronized batch normalization across devices to ensure consistent feature scaling.

\subsection{Baselines and Metrics}
We grouped baselines into three categories: 
(1) unimodal foundations for pathological images, 
(2) unimodal foundations for transcriptomics, and 
(3) multimodal contrastive learning frameworks.

\noindent \textbf{Pathology Baselines:}  
All pathology foundation baselines (P.) served as frozen encoders to embed pathological images of individual ST spots, which were then used in downstream tasks.
The baselines include CTransPath\cite{wang2022transformer}, UNI\cite{chen2024uni}, Prov-GigaPath\cite{xu2024gigapath}, and Hibou\cite{nechaev2024hibou}, all trained on large-scale WSIs. Additionally, CONCH\cite{lu2024visual} was a visual-language foundational model trained on paired historical images and medical report texts.

\noindent \textbf{Transcriptomic Baselines:}
Transcriptomic foundation baselines (G.) were applied to extract transcriptomic features from each ST spot, similar to the pathology baselines. 
The baselines include scFoundation\cite{hao2024large} and scGPT\cite{cui2023scGPT}, recent foundation models pretrained on large-scale single-cell RNA sequencing data. 
We also included Scanpy\cite{wolf2018scanpy}, the most prevalent toolkit for transcriptomic data analysis.

\noindent \textbf{Multimodal Baselines:}
We also pretrained popular multimodal contrastive learning frameworks, CLIP\cite{radford2021learning} and PLIP\cite{huang2023visual}, as baselines.~Following the approach in STimage-1K4M\cite{chen2024stimage1k4m}, we used a fully connected (FC) layer to compress the original gene expression profile into a 32-dimensional embedding. 
Simultaneously, a pretrained image encoder was used, followed by an FC layer that projected images into a 32-dimensional representation. 
We loaded pretrained parameters (ViT-B/32) for CLIP from \textit{openai/clip-vit-base-patch32}, and for PLIP, pretrained parameters (ViT-L/14) from \textit{vinid/plip} on Hugging Face. 
Hyperparameters were chosen to match those used for CLIP training.


\noindent \textbf{Metrics:}
The performance of ST-Align and other foundation model in two downstream tasks were evaluated through (1) the adjusted rand index (ARI, higher is better), which measure the similarity between true region and clusters based on embeddings, and 
(2) mean-square error (MSE, lower is better) that indicate the deviation  between predicting gene expression and true expression level among all spots.

\begin{table}[h!]
    \small
    \centering
    \caption{\textbf{Ablation Study.} Concatenate denotes the equal fusion of image and gene features. ABFN, \(\mathcal{L}_{NS}\), and AE represent our three designs: Attention-Based Fusion Network, niche-spot loss, and trainable encoder, respectively. Note that since scGPT and Concatenate incorporate genetic information, we did not conduct experiments on them for gene prediction.}
    \begin{tabular}{cc|c}
        \toprule
        \multirow{2}{*}{Model} & \multirow{2}{*}{\makecell[c]{Clustering ARI $\uparrow$}} &  \multirow{2}{*}{\makecell[c]{Prediction MSE $\downarrow$}} \\
        &  &  \\
        \midrule
        UNI & 0.1089 & 0.2014 \\
        scGPT & 0.2615 & - \\
        \hline
        Concatenate & 0.1106 & - \\  
        ABFN $+$ \(\mathcal{L}_{NS}\) & 0.2590 & 0.1801 \\ 
        AE $+$ ABFN & 0.1620 & 0.1710 \\ 
        AE $+$ \(\mathcal{L}_{NS}\) $+$ ABFN & 0.3396 & 0.1682 \\
        \bottomrule
    \end{tabular}
    \label{table:performance_comparison}
\end{table}
\renewcommand{\arraystretch}{1.2}
\begin{table*}[ht]
\centering
\caption{
\textbf{Performance of predicting gene expression based on images.} Best performance in \textbf{bold}, second best \underline{underlined}. The standard deviation is reported over six datasets and evaluated using the MSE metric.
}
\label{tab:results-two}
\setlength\tabcolsep{2.5pt}
\resizebox{\linewidth}{!}{
\begin{tabular}{@{}lcccccccccccc@{}}
\toprule
Model & \multicolumn{3}{c}{Layer Marker Genes} & \multicolumn{3}{c}{Laminar} & \multicolumn{3}{c}{Non-Laminar} & Overall \\     
\cmidrule(lr){2-4}\cmidrule(lr){5-7}\cmidrule(lr){8-10}
& FABP7 & CCK & PVALB & PCP4 & MOBP & SNAP25 & IGKC & HBB & NPY & \\ 
\midrule
CTranPath & 0.4645\scriptsize± 0.105 & 0.2002\scriptsize± 0.060 & 0.1667\scriptsize± 0.072 & 0.1590\scriptsize± 0.099 & 0.2119\scriptsize± 0.118 & 0.3632\scriptsize± 0.094 & 0.0820\scriptsize± 0.042 & \underline{0.0583}\scriptsize± 0.026 & 0.0315\scriptsize± 0.034 & 0.1927 \\ 

CONCH & 0.4396\scriptsize± 0.131 & \textbf{0.1680}\scriptsize± 0.069 & 0.1749\scriptsize± 0.067 & 0.1890\scriptsize± 0.122 & 0.2222\scriptsize± 0.151 & 0.3471\scriptsize± 0.091 & 0.0672\scriptsize± 0.033 & 0.0891\scriptsize± 0.048 & \underline{0.0271}\scriptsize± 0.010 & 0.1916 \\ 

Prov-GigaPath & 0.4309\scriptsize± 0.081 & 0.2105\scriptsize± 0.078 & 0.2050\scriptsize± 0.118 & 0.1611\scriptsize± 0.077 & 0.2595\scriptsize± 0.167 & 0.3804\scriptsize± 0.123 & \underline{0.0582}\scriptsize± 0.014 & 0.0720\scriptsize± 0.024 & 0.0385\scriptsize± 0.047 & 0.2018 \\ 

Hibou & 0.4056\scriptsize± 0.091 & 0.1842\scriptsize± 0.082 & 0.2042\scriptsize± 0.076 & 0.1729\scriptsize± 0.102 & 0.2219\scriptsize± 0.1382 & \underline{0.3065}\scriptsize± 0.085 & 0.0746\scriptsize± 0.021 & 0.0656\scriptsize± 0.031 & 0.0278\scriptsize± 0.008 & 0.1848 \\ 

UNI & 0.4782\scriptsize± 0.101 & 0.1943\scriptsize± 0.049 & 0.1824\scriptsize± 0.056 & 0.1508\scriptsize± 0.088 & 0.2831\scriptsize± 0.200 & 0.3834\scriptsize± 0.070 & 0.0692\scriptsize± 0.045 & \textbf{0.0494}\scriptsize± 0.021 & 0.0274\scriptsize± 0.024 & 0.2014 \\ 
\hline

CLIP & \underline{0.3944}\scriptsize± 0.106 & 0.1966\scriptsize± 0.088 & 0.1703\scriptsize± 0.068 & 0.1559\scriptsize± 0.090 & \underline{0.2061}\scriptsize± 0.083 & 0.3205\scriptsize± 0.112 & 0.0758\scriptsize± 0.038 & 0.1118\scriptsize± 0.030 & 0.0344\scriptsize± 0.040 & \underline{0.1840} \\ 

PLIP & 0.3951\scriptsize± 0.106 & 0.1936\scriptsize± 0.090 & \underline{0.1650}\scriptsize± 0.069 & \underline{0.1502}\scriptsize± 0.089 & 0.2064\scriptsize± 0.080 & 0.3230\scriptsize± 0.110 & 0.0753\scriptsize± 0.038 & 0.1257\scriptsize± 0.042 & 0.0344\scriptsize± 0.039 & 0.1854 \\ 

\textbf{ST-Align} & \textbf{0.3824}\scriptsize± 0.075 & \underline{0.1754}\scriptsize± 0.079 & \textbf{0.1644}\scriptsize± 0.087 & \textbf{0.1480}\scriptsize± 0.083 & \textbf{0.1898}\scriptsize± 0.113 & \textbf{0.2982}\scriptsize± 0.061 & \textbf{0.0547}\scriptsize± 0.031 & 0.0743\scriptsize± 0.022 & \textbf{0.0269}\scriptsize± 0.034 & \textbf{0.1682} \\ 

\bottomrule
\end{tabular}
}
\end{table*}

\subsection{Spatial Clustering Identification}
ST is commonly used to explore spatial regions within tissue slices. 
Here, we evaluated ST-Align and baseline models in identifying spatial regions by testing on six independent human brain slices from~\cite{maynard2021transcriptome}.
Table~\ref{tab:results-one} shows that ST-Align achieved the best overall performance, outperforming both (1) unimodal foundation models and (2) multimodal baselines in a zero-shot setting.


\noindent \textbf{ST-Align vs. Unimodal:} 
As showed in Table~\ref{tab:results-one}, ST-Align outperformed all unimodal foundation baselines across all test slices. 
Genetic foundation models exceeded pathological models by +15.49\%, indicating that relying solely on pathological images without considering genetic information is insufficient for accurate spatial domain identification. 
Notably, ST-Align achieved improvements of +23.22\% and +7.73\% over pathological and genetic foundation models, respectively. 
Although CLIP and PLIP performed comparably to genetic foundation models, their performance was limited by using only a simple MLP for gene modeling. 
These results highlight the substantial benefits of integrating genetic and morphological features for distinguishing biological structures.


\noindent \textbf{ST-Align vs. Multimodal:} 
Comparing ST-Align to popular multimodal frameworks CLIP and PLIP, ST-Align achieved +8.35\% and +9.13\% higher ARI scores, respectively. 
These results demonstrate ST-Align's effectiveness in leveraging the ABFN and a two-stage contrastive learning approach for modeling ST data.



\subsection{Spot Gene Expression Prediction}
Predicting gene expression at the single-spot level can potentially reduce the need for costly and time-consuming library preparation and sequencing. 
In this experiment, we used the image encoders from ST-Align and other baseline models (excluding genetic foundation models) in cooperating with an MLP, trained on 80\% of the spots to predict gene expression values for the remaining spots. 
The prediction results for nine genes, categorized into three groups, are presented in Table~\ref{tab:results-two}.

\noindent \textbf{Unimodal vs. Multimodal:}
Compared to unimodal methods, the multimodal model pretrained on other ST datasets achieved better results overall. 
The multimodal models showed performance improvements of \(+9.26\%\) and \(+12.64\%\) in predicting Layer Marker Genes and Laminar Genes, respectively, but a decrease of \(-21.99\%\) for Non-Laminar Genes, while ST-Align achieved a \(+6.97\%\) improvement in Non-Laminar Genes. 
Unlike Layer Marker and Laminar Genes, Non-Laminar Genes are not structure-specific. 
The observed contrasting performance between ST-Align and the baselines underscores the importance of approaches of incorporating genetic features during pre-training phase.


\noindent \textbf{ST-Align vs. Multimodal:} 
Compared to other multimodal methods, ST-Align achieved performance improvements of \(+3.16\%\), \(+4.51\%\), and \(+23.74\%\) in predicting Layer Marker Genes, Laminar Genes, and Non-Laminar Genes, respectively, with the largest gain observed in Non-Laminar Genes. 
These results highlight the necessity of the ABFN and AEs and the spatial perception module. 
In summary, ST-Align serves as an effective method for multimodal joint analysis and gene expreesion prediction.



\subsection{Ablation Study}
To evaluate the modules in ST-Align, we performed a series of ablation studies on two downstream tasks, with the results displayed in Table~\ref{table:performance_comparison}.

\begin{figure*}[t]
  \centering
   \includegraphics[width=17cm,height=8cm]{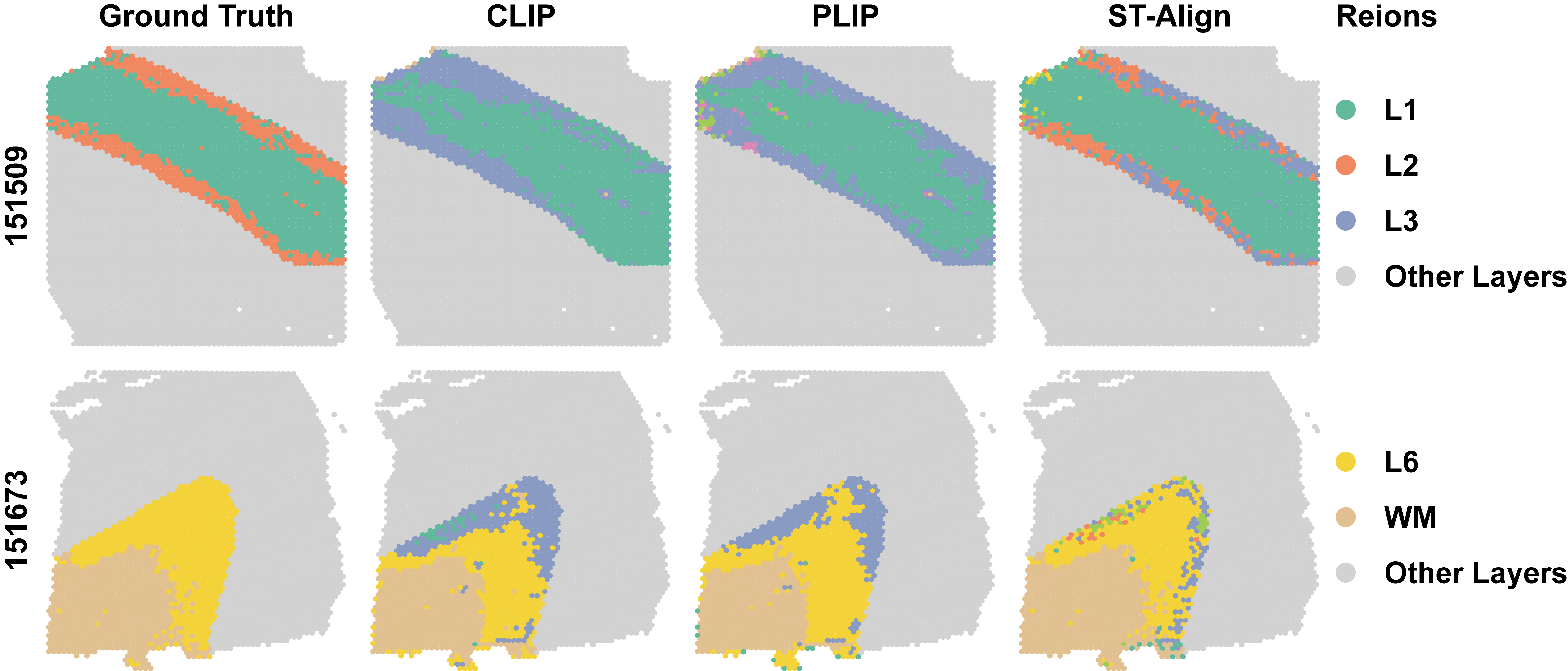}
   \caption{\textbf{Zero-shot Spatial Clustering Results.}  
   The performance of methods in identifying spatial domains was evaluated by comparing ST-Align with existing methods CLIP and PLIP, using human annotation as the ground truth. Each row represents distinct slices (151509 and 151673) derived from different samples. Each color corresponds to a distinct spatial region, ranging from WM (White Matter) to L1.}
   \label{fig:visua_task}
\end{figure*}

\noindent \textbf{AEs and ABFN:}
ST-Align utilizes AEs and ABFN to capture and fuse domain-specific knowledge with ST-specific information efficiently.
First, we ablated the AEs, resulting in a performance reduction of 8.06\% and 6.61\% in the two tasks, respectively. 
To furtherly investigate the strategy of ST-Align for modeling ST data, we replaced the ABFN$+$AE combination with direct concatenation of unimodal embeddings, which reduced performance by 5.14\%  in the first downstream task. 
These results underscore the effectiveness of AEs and ABFN in modeling and integrating ST-specific pathological image and genetic data.

\noindent \textbf{Spot-Niche contrastive learning:}
We further ablated the spot-niche contrastive loss \(\mathcal{L}_{NS}\), which guides alignment between individual spots and their corresponding niches. 
Results indicate that incorporating \(\mathcal{L}_{NS}\) improves performance of ST-Align by \(+\)17.76\% and \(+\)1.64\% in the two tasks, respectively. 
Comparing ABFN$+$\(\mathcal{L}_{NS}\) with ABFN$+$AE, we observed improved performance in the spatial identification task but a reduction in the gene prediction task. 
This finding suggests that \(\mathcal{L}_{NS}\) likely enhances the ability of ST-Align to model spatial relationships between fine-grained and coarse-grained data, while ABFN$+$AE is more effective at capturing intrinsic characteristics within ST data.

\subsection{Visualization}
To attribute the effectiveness of multimodal strategies in identifying spatial clusters, we visualized the predicted cluster labels of ST-Align, CLIP, and PLIP in a zero-shot setting. 
As shown in Figure~\ref{fig:visua_task}, in slice 151509, layers L1 and L2 are continuous but display subtle differences in structure that CLIP and PLIP fail to distinguish accurately, whereas ST-Align successfully differentiates them.
Additionally, in slice 151673, ST-Align more effectively delineates the boundary between the white matter (WM) and layer L6 compared to CLIP and PLIP.
\section{Conclusions}
\label{sec:Conclusions}
In this paper, we introduced ST-Align, the first multimodal foundation model for ST. 
ST-Align was pretrained on 1.3 million spots with corresponding niche data from 573 human tissue slices, encompassing normal, diseased, and cancerous status.
Overall, ST-Align significantly outperforms all baseline models across two downstream tasks: spatial domain identification and gene expression prediction. 
These results emphasize the potential of tailored modules for effectively modeling the unique pathological image and genetic features in ST data.
Future work includes implementing stricter data quality control and expanding to incorporate more data and additional modalities to enhance versatility. 
Additionally, exploring ST in further applications, such as differentiating niches associated with clinical phenotypes, presents promising research directions.
{
    \small
    \bibliographystyle{ieeenat_fullname}
    \bibliography{main}

\begin{thebibliography}{58}
\providecommand{\natexlab}[1]{#1}
\providecommand{\url}[1]{\texttt{#1}}
\expandafter\ifx\csname urlstyle\endcsname\relax
  \providecommand{\doi}[1]{doi: #1}\else
  \providecommand{\doi}{doi: \begingroup \urlstyle{rm}\Url}\fi

\bibitem[Bejarano et~al.(2021)Bejarano, Jord{\=a}o, and Joyce]{bejarano2021therapeutic}
Leire Bejarano, Marta~JC Jord{\=a}o, and Johanna~A Joyce.
\newblock Therapeutic targeting of the tumor microenvironment.
\newblock \emph{Cancer discovery}, 11\penalty0 (4):\penalty0 933--959, 2021.

\bibitem[Benjamin et~al.(2024)Benjamin, Bhandari, Kepple, Qi, Shang, Xing, An, Zhang, Hou, Crockford, et~al.]{benjamin2024multiscale}
Katherine Benjamin, Aneesha Bhandari, Jessica~D Kepple, Rui Qi, Zhouchun Shang, Yanan Xing, Yanru An, Nannan Zhang, Yong Hou, Tanya~L Crockford, et~al.
\newblock Multiscale topology classifies cells in subcellular spatial transcriptomics.
\newblock \emph{Nature}, pages 1--7, 2024.

\bibitem[Bian et~al.(2024)Bian, Chen, Dong, Li, Hao, Chen, Hu, Sun, Wei, and Zhang]{bian2024scmulan}
Haiyang Bian, Yixin Chen, Xiaomin Dong, Chen Li, Minsheng Hao, Sijie Chen, Jinyi Hu, Maosong Sun, Lei Wei, and Xuegong Zhang.
\newblock scmulan: a multitask generative pre-trained language model for single-cell analysis.
\newblock In \emph{International Conference on Research in Computational Molecular Biology}, pages 479--482. Springer, 2024.

\bibitem[Chen et~al.(2024{\natexlab{a}})Chen, Zhou, Wu, Zhang, Li, and Li]{chen2024stimage1k4m}
Jiawen Chen, Muqing Zhou, Wenrong Wu, Jinwei Zhang, Yun Li, and Didong Li.
\newblock Stimage-1k4m: A histopathology image-gene expression dataset for spatial transcriptomics, 2024{\natexlab{a}}.

\bibitem[Chen et~al.(2015)Chen, Boettiger, Moffitt, Wang, and Zhuang]{chen2015spatially}
Kok~Hao Chen, Alistair~N Boettiger, Jeffrey~R Moffitt, Siyuan Wang, and Xiaowei Zhuang.
\newblock Spatially resolved, highly multiplexed rna profiling in single cells.
\newblock \emph{Science}, 348\penalty0 (6233):\penalty0 aaa6090, 2015.

\bibitem[Chen et~al.(2023)Chen, Ding, Lu, Williamson, Jaume, Chen, Zhang, Shao, Song, Shaban, et~al.]{chen2023general}
Richard~J Chen, Tong Ding, Ming~Y Lu, Drew~FK Williamson, Guillaume Jaume, Bowen Chen, Andrew Zhang, Daniel Shao, Andrew~H Song, Muhammad Shaban, et~al.
\newblock A general-purpose self-supervised model for computational pathology.
\newblock \emph{arXiv preprint arXiv:2308.15474}, 2023.

\bibitem[Chen et~al.(2024{\natexlab{b}})Chen, Ding, Lu, Williamson, Jaume, Chen, Zhang, Shao, Song, Shaban, et~al.]{chen2024uni}
Richard~J Chen, Tong Ding, Ming~Y Lu, Drew~FK Williamson, Guillaume Jaume, Bowen Chen, Andrew Zhang, Daniel Shao, Andrew~H Song, Muhammad Shaban, et~al.
\newblock Towards a general-purpose foundation model for computational pathology.
\newblock \emph{Nature Medicine}, 2024{\natexlab{b}}.

\bibitem[Chen et~al.(2024{\natexlab{c}})Chen, Xie, Lin, Song, Yang, and Yu]{chen2024survmamba}
Ying Chen, Jiajing Xie, Yuxiang Lin, Yuhang Song, Wenxian Yang, and Rongshan Yu.
\newblock Survmamba: State space model with multi-grained multi-modal interaction for survival prediction.
\newblock \emph{arXiv preprint arXiv:2404.08027}, 2024{\natexlab{c}}.

\bibitem[Christensen et~al.(2024)Christensen, Vukadinovic, Yuan, and Ouyang]{christensen2024vision}
Matthew Christensen, Milos Vukadinovic, Neal Yuan, and David Ouyang.
\newblock Vision--language foundation model for echocardiogram interpretation.
\newblock \emph{Nature Medicine}, pages 1--8, 2024.

\bibitem[Cui et~al.(2024)Cui, Wang, Maan, Pang, Luo, Duan, and Wang]{cui2023scGPT}
Haotian Cui, Chloe Wang, Hassaan Maan, Kuan Pang, Fengning Luo, Nan Duan, and Bo Wang.
\newblock scgpt: toward building a foundation model for single-cell multi-omics using generative ai.
\newblock \emph{Nature Methods}, pages 1--11, 2024.

\bibitem[De~Visser and Joyce(2023)]{de2023evolving}
Karin~E De~Visser and Johanna~A Joyce.
\newblock The evolving tumor microenvironment: From cancer initiation to metastatic outgrowth.
\newblock \emph{Cancer cell}, 41\penalty0 (3):\penalty0 374--403, 2023.

\bibitem[Ding et~al.(2023)Ding, Zhou, Metaxas, and Zhang]{ding2023pathology}
Kexin Ding, Mu Zhou, Dimitris~N Metaxas, and Shaoting Zhang.
\newblock Pathology-and-genomics multimodal transformer for survival outcome prediction.
\newblock In \emph{International Conference on Medical Image Computing and Computer-Assisted Intervention}, pages 622--631. Springer, 2023.

\bibitem[Hao et~al.(2024)Hao, Gong, Zeng, Liu, Guo, Cheng, Wang, Ma, Zhang, and Song]{hao2024large}
Minsheng Hao, Jing Gong, Xin Zeng, Chiming Liu, Yucheng Guo, Xingyi Cheng, Taifeng Wang, Jianzhu Ma, Xuegong Zhang, and Le Song.
\newblock Large-scale foundation model on single-cell transcriptomics.
\newblock \emph{Nature Methods}, pages 1--11, 2024.

\bibitem[He et~al.(2015)He, Zhang, Ren, and Sun]{He2015}
Kaiming He, Xiangyu Zhang, Shaoqing Ren, and Jian Sun.
\newblock Deep residual learning for image recognition.
\newblock \emph{arXiv preprint arXiv:1512.03385}, 2015.

\bibitem[Hu et~al.(2021)Hu, Li, Coleman, Schroeder, Ma, Irwin, Lee, Shinohara, and Li]{hu2021spagcn}
Jian Hu, Xiangjie Li, Kyle Coleman, Amelia Schroeder, Nan Ma, David~J Irwin, Edward~B Lee, Russell~T Shinohara, and Mingyao Li.
\newblock Spagcn: Integrating gene expression, spatial location and histology to identify spatial domains and spatially variable genes by graph convolutional network.
\newblock \emph{Nature methods}, 18\penalty0 (11):\penalty0 1342--1351, 2021.

\bibitem[Hu et~al.(2024)Hu, Rong, Xu, Xie, Peng, Gao, and Tan]{hu2024unsupervised}
Yuxuan Hu, Jiazhen Rong, Yafei Xu, Runzhi Xie, Jacqueline Peng, Lin Gao, and Kai Tan.
\newblock Unsupervised and supervised discovery of tissue cellular neighborhoods from cell phenotypes.
\newblock \emph{Nature Methods}, 21\penalty0 (2):\penalty0 267--278, 2024.

\bibitem[Huang et~al.(2023)Huang, Bianchi, Yuksekgonul, Montine, and Zou]{huang2023visual}
Zhi Huang, Federico Bianchi, Mert Yuksekgonul, Thomas~J Montine, and James Zou.
\newblock A visual--language foundation model for pathology image analysis using medical twitter.
\newblock \emph{Nature medicine}, 29\penalty0 (9):\penalty0 2307--2316, 2023.

\bibitem[Ikezogwo et~al.(2023)Ikezogwo, Seyfioglu, Ghezloo, Geva, Sheikh~Mohammed, Anand, Krishna, and Shapiro]{ikezogwo2024quilt}
Wisdom Ikezogwo, Saygin Seyfioglu, Fatemeh Ghezloo, Dylan Geva, Fatwir Sheikh~Mohammed, Pavan~Kumar Anand, Ranjay Krishna, and Linda Shapiro.
\newblock Quilt-1m: One million image-text pairs for histopathology.
\newblock In \emph{Advances in Neural Information Processing Systems}, pages 37995--38017. Curran Associates, Inc., 2023.

\bibitem[Jaume et~al.(2024{\natexlab{a}})Jaume, Doucet, Song, Lu, Almagro-P{\'e}rez, Wagner, Vaidya, Chen, Williamson, Kim, et~al.]{jaume2024hest}
Guillaume Jaume, Paul Doucet, Andrew~H Song, Ming~Y Lu, Cristina Almagro-P{\'e}rez, Sophia~J Wagner, Anurag~J Vaidya, Richard~J Chen, Drew~FK Williamson, Ahrong Kim, et~al.
\newblock Hest-1k: A dataset for spatial transcriptomics and histology image analysis.
\newblock \emph{arXiv preprint arXiv:2406.16192}, 2024{\natexlab{a}}.

\bibitem[Jaume et~al.(2024{\natexlab{b}})Jaume, Oldenburg, Vaidya, Chen, Williamson, Peeters, Song, and Mahmood]{jaume2024transcriptomics}
Guillaume Jaume, Lukas Oldenburg, Anurag Vaidya, Richard~J Chen, Drew~FK Williamson, Thomas Peeters, Andrew~H Song, and Faisal Mahmood.
\newblock Transcriptomics-guided slide representation learning in computational pathology.
\newblock In \emph{Proceedings of the IEEE/CVF Conference on Computer Vision and Pattern Recognition}, pages 9632--9644, 2024{\natexlab{b}}.

\bibitem[Jaume et~al.(2024{\natexlab{c}})Jaume, Vaidya, Chen, Williamson, Liang, and Mahmood]{jaume2024modeling}
Guillaume Jaume, Anurag Vaidya, Richard~J Chen, Drew~FK Williamson, Paul~Pu Liang, and Faisal Mahmood.
\newblock Modeling dense multimodal interactions between biological pathways and histology for survival prediction.
\newblock In \emph{Proceedings of the IEEE/CVF Conference on Computer Vision and Pattern Recognition}, pages 11579--11590, 2024{\natexlab{c}}.

\bibitem[Jaume et~al.(2024{\natexlab{d}})Jaume, Vaidya, Zhang, Song, Chen, Sahai, Mo, Madrigal, Le, and Mahmood]{jaume2024multistain}
Guillaume Jaume, Anurag Vaidya, Andrew Zhang, Andrew~H Song, Richard~J Chen, Sharifa Sahai, Dandan Mo, Emilio Madrigal, Long~Phi Le, and Faisal Mahmood.
\newblock Multistain pretraining for slide representation learning in pathology.
\newblock \emph{arXiv preprint arXiv:2408.02859}, 2024{\natexlab{d}}.

\bibitem[Khwaja et~al.(2024)Khwaja, Song, Agarunov, and Huang]{khwaja2024celle}
Emaad Khwaja, Yun Song, Aaron Agarunov, and Bo Huang.
\newblock Celle-2: Translating proteins to pictures and back with a bidirectional text-to-image transformer.
\newblock \emph{Advances in Neural Information Processing Systems}, 36, 2024.

\bibitem[Li et~al.(2024)Li, Chen, Chen, Yu, Yang, Wang, Ding, and Han]{Li_2024_CVPR}
Hao Li, Ying Chen, Yifei Chen, Rongshan Yu, Wenxian Yang, Liansheng Wang, Bowen Ding, and Yuchen Han.
\newblock Generalizable whole slide image classification with fine-grained visual-semantic interaction.
\newblock In \emph{Proceedings of the IEEE/CVF Conference on Computer Vision and Pattern Recognition (CVPR)}, pages 11398--11407, 2024.

\bibitem[Lu et~al.(2024)Lu, Chen, Williamson, Chen, Liang, Ding, Jaume, Odintsov, Le, Gerber, et~al.]{lu2024visual}
Ming~Y Lu, Bowen Chen, Drew~FK Williamson, Richard~J Chen, Ivy Liang, Tong Ding, Guillaume Jaume, Igor Odintsov, Long~Phi Le, Georg Gerber, et~al.
\newblock A visual-language foundation model for computational pathology.
\newblock \emph{Nature Medicine}, 30\penalty0 (3):\penalty0 863--874, 2024.

\bibitem[Ma and Zhou(2024)]{ma2024accurate}
Ying Ma and Xiang Zhou.
\newblock Accurate and efficient integrative reference-informed spatial domain detection for spatial transcriptomics.
\newblock \emph{Nature Methods}, pages 1--14, 2024.

\bibitem[Maynard et~al.(2021)Maynard, Collado-Torres, Weber, Uytingco, Barry, Williams, Catallini, Tran, Besich, Tippani, et~al.]{maynard2021transcriptome}
Kristen~R Maynard, Leonardo Collado-Torres, Lukas~M Weber, Cedric Uytingco, Brianna~K Barry, Stephen~R Williams, Joseph~L Catallini, Matthew~N Tran, Zachary Besich, Madhavi Tippani, et~al.
\newblock Transcriptome-scale spatial gene expression in the human dorsolateral prefrontal cortex.
\newblock \emph{Nature neuroscience}, 24\penalty0 (3):\penalty0 425--436, 2021.

\bibitem[Moses and Pachter(2022)]{moses2022museum}
Lambda Moses and Lior Pachter.
\newblock Museum of spatial transcriptomics.
\newblock \emph{Nature methods}, 19\penalty0 (5):\penalty0 534--546, 2022.

\bibitem[Nechaev et~al.(2024)Nechaev, Pchelnikov, and Ivanova]{nechaev2024hibou}
Dmitry Nechaev, Alexey Pchelnikov, and Ekaterina Ivanova.
\newblock Hibou: A family of foundational vision transformers for pathology.
\newblock \emph{arXiv preprint arXiv:2406.05074}, 2024.

\bibitem[Niazi et~al.(2019)Niazi, Parwani, and Gurcan]{niazi2019digital}
Muhammad Khalid~Khan Niazi, Anil~V Parwani, and Metin~N Gurcan.
\newblock Digital pathology and artificial intelligence.
\newblock \emph{The lancet oncology}, 20\penalty0 (5):\penalty0 e253--e261, 2019.

\bibitem[Radford et~al.(2021)Radford, Kim, Hallacy, Ramesh, Goh, Agarwal, Sastry, Askell, Mishkin, et~al.]{radford2021learning}
Alec Radford, Jong~Wook Kim, Chris Hallacy, Aditya Ramesh, Gabriel Goh, Sandhini Agarwal, Girish Sastry, Amanda Askell, Pamela Mishkin, et~al.
\newblock Learning transferable visual models from natural language supervision.
\newblock In \emph{Proceedings of the International Conference on Machine Learning}, 2021.

\bibitem[Schuhmann et~al.(2022)Schuhmann, Beaumont, Vencu, Gordon, Wightman, Cherti, Coombes, Katta, Mullis, Wortsman, et~al.]{schuhmann2022laion}
Christoph Schuhmann, Romain Beaumont, Richard Vencu, Cade Gordon, Ross Wightman, Mehdi Cherti, Theo Coombes, Aarush Katta, Clayton Mullis, Mitchell Wortsman, et~al.
\newblock Laion-5b: An open large-scale dataset for training next generation image-text models.
\newblock \emph{Advances in Neural Information Processing Systems}, 35:\penalty0 25278--25294, 2022.

\bibitem[Si et~al.(2024)Si, Lee, Hwang, Yun, Cheng, Cho, Quiros, Nusrat, Zhang, Jun, et~al.]{si2024ficture}
Yichen Si, ChangHee Lee, Yongha Hwang, Jeong~H Yun, Weiqiu Cheng, Chun-Seok Cho, Miguel Quiros, Asma Nusrat, Weizhou Zhang, Goo Jun, et~al.
\newblock Ficture: scalable segmentation-free analysis of submicron-resolution spatial transcriptomics.
\newblock \emph{Nature Methods}, pages 1--12, 2024.

\bibitem[Song et~al.(2024)Song, Chen, Ding, Williamson, Jaume, and Mahmood]{song2024morphological}
Andrew~H Song, Richard~J Chen, Tong Ding, Drew~FK Williamson, Guillaume Jaume, and Faisal Mahmood.
\newblock Morphological prototyping for unsupervised slide representation learning in computational pathology.
\newblock In \emph{Proceedings of the IEEE/CVF Conference on Computer Vision and Pattern Recognition}, pages 11566--11578, 2024.

\bibitem[Sun et~al.(2024)Sun, Zhu, Zheng, Zhang, Sun, Shui, Zhang, Li, and Yang]{sun2024pathasst}
Yuxuan Sun, Chenglu Zhu, Sunyi Zheng, Kai Zhang, Lin Sun, Zhongyi Shui, Yunlong Zhang, Honglin Li, and Lin Yang.
\newblock Pathasst: A generative foundation ai assistant towards artificial general intelligence of pathology.
\newblock In \emph{Proceedings of the AAAI Conference on Artificial Intelligence}, pages 5034--5042, 2024.

\bibitem[Sun et~al.(2025)Sun, Wu, Zhu, Zheng, Chen, Zhang, Zhang, Wan, Lan, Zheng, et~al.]{sun2025pathmmu}
Yuxuan Sun, Hao Wu, Chenglu Zhu, Sunyi Zheng, Qizi Chen, Kai Zhang, Yunlong Zhang, Dan Wan, Xiaoxiao Lan, Mengyue Zheng, et~al.
\newblock Pathmmu: A massive multimodal expert-level benchmark for understanding and reasoning in pathology.
\newblock In \emph{European Conference on Computer Vision}, pages 56--73. Springer, 2025.

\bibitem[Tang et~al.(2024)Tang, Zhou, Huang, Zhu, Zhang, and Liu]{tang2024feature}
Wenhao Tang, Fengtao Zhou, Sheng Huang, Xiang Zhu, Yi Zhang, and Bo Liu.
\newblock Feature re-embedding: Towards foundation model-level performance in computational pathology.
\newblock In \emph{Proceedings of the IEEE/CVF Conference on Computer Vision and Pattern Recognition}, pages 11343--11352, 2024.

\bibitem[Tian et~al.(2024)Tian, Liu, Wei, Fu, Sun, Liu, Sui, Tian, Nemeth, Feng, et~al.]{tian2024prediction}
Fei Tian, Dong Liu, Na Wei, Qianqian Fu, Lin Sun, Wei Liu, Xiaolong Sui, Kathryn Tian, Genevieve Nemeth, Jingyu Feng, et~al.
\newblock Prediction of tumor origin in cancers of unknown primary origin with cytology-based deep learning.
\newblock \emph{Nature Medicine}, pages 1--11, 2024.

\bibitem[Tong et~al.(2023)Tong, Lin, Yang, Song, Zhang, Xie, Tian, Luo, Liang, Huang, et~al.]{tong2023prioritizing}
Mengsha Tong, Yuxiang Lin, Wenxian Yang, Jinsheng Song, Zheyang Zhang, Jiajing Xie, Jingyi Tian, Shijie Luo, Chenyu Liang, Jialiang Huang, et~al.
\newblock Prioritizing prognostic-associated subpopulations and individualized recurrence risk signatures from single-cell transcriptomes of colorectal cancer.
\newblock \emph{Briefings in Bioinformatics}, 24\penalty0 (3):\penalty0 bbad078, 2023.

\bibitem[Vaswani et~al.(2017)Vaswani, Shazeer, Parmar, Uszkoreit, Jones, Gomez, Kaiser, and Polosukhin]{vaswani2017attention}
Ashish Vaswani, Noam Shazeer, Niki Parmar, Jakob Uszkoreit, Llion Jones, Aidan~N. Gomez, Łukasz Kaiser, and Iliya Polosukhin.
\newblock Attention is all you need.
\newblock In \emph{Advances in Neural Information Processing Systems}. Curran Associates, Inc., 2017.

\bibitem[Wang et~al.(2024{\natexlab{a}})Wang, Wu, Xiong, Qu, Fang, and Bao]{wang2024crost}
Guoliang Wang, Song Wu, Zhuang Xiong, Hongzhu Qu, Xiangdong Fang, and Yiming Bao.
\newblock Crost: a comprehensive repository of spatial transcriptomics.
\newblock \emph{Nucleic Acids Research}, 52\penalty0 (D1):\penalty0 D882--D890, 2024{\natexlab{a}}.

\bibitem[Wang et~al.(2018)Wang, Allen, Wright, Sylwestrak, Samusik, Vesuna, Evans, Liu, Ramakrishnan, Liu, et~al.]{wang2018three}
Xiao Wang, William~E Allen, Matthew~A Wright, Emily~L Sylwestrak, Nikolay Samusik, Sam Vesuna, Kathryn Evans, Cindy Liu, Charu Ramakrishnan, Jia Liu, et~al.
\newblock Three-dimensional intact-tissue sequencing of single-cell transcriptional states.
\newblock \emph{Science}, 361\penalty0 (6400):\penalty0 eaat5691, 2018.

\bibitem[Wang et~al.(2022{\natexlab{a}})Wang, Yang, Zhang, Wang, Zhang, Yang, Huang, and Han]{wang2022transformer}
Xiyue Wang, Sen Yang, Jun Zhang, Minghui Wang, Jing Zhang, Wei Yang, Junzhou Huang, and Xiao Han.
\newblock Transformer-based unsupervised contrastive learning for histopathological image classification.
\newblock \emph{Medical image analysis}, 81:\penalty0 102559, 2022{\natexlab{a}}.

\bibitem[Wang et~al.(2024{\natexlab{b}})Wang, Zhao, Marostica, Yuan, Jin, Zhang, Li, Tang, Wang, Li, et~al.]{wang2024pathology}
Xiyue Wang, Junhan Zhao, Eliana Marostica, Wei Yuan, Jietian Jin, Jiayu Zhang, Ruijiang Li, Hongping Tang, Kanran Wang, Yu Li, et~al.
\newblock A pathology foundation model for cancer diagnosis and prognosis prediction.
\newblock \emph{Nature}, pages 1--9, 2024{\natexlab{b}}.

\bibitem[Wang et~al.(2022{\natexlab{b}})Wang, Song, Wang, Chen, Xie, Xiao, Wang, and Wang]{wang2022sprod}
Yunguan Wang, Bing Song, Shidan Wang, Mingyi Chen, Yang Xie, Guanghua Xiao, Li Wang, and Tao Wang.
\newblock Sprod for de-noising spatially resolved transcriptomics data based on position and image information.
\newblock \emph{Nature methods}, 19\penalty0 (8):\penalty0 950--958, 2022{\natexlab{b}}.

\bibitem[Wen et~al.(2024)Wen, Tang, Dai, Ding, Jin, Xie, and Tang]{2024wencellplm}
Hongzhi Wen, Wenzhuo Tang, Xinnan Dai, Jiayuan Ding, Wei Jin, Yuying Xie, and Jiliang Tang.
\newblock Cellplm: Pre-training of cell language model beyond single cells.
\newblock In \emph{The Twelfth International Conference on Learning Representations}, 2024.

\bibitem[Wolf et~al.(2018)Wolf, Angerer, and Theis]{wolf2018scanpy}
F.~Alexander Wolf, Philipp Angerer, and Fabian~J. Theis.
\newblock Scanpy: large-scale single-cell gene expression data analysis.
\newblock \emph{Genome Biology}, 2018.

\bibitem[Xiong et~al.(2024)Xiong, Chen, Zheng, Wei, Zheng, Sung, and King]{xiong2024mome}
Conghao Xiong, Hao Chen, Hao Zheng, Dong Wei, Yefeng Zheng, Joseph~JY Sung, and Irwin King.
\newblock Mome: Mixture of multimodal experts for cancer survival prediction.
\newblock In \emph{International Conference on Medical Image Computing and Computer-Assisted Intervention}, pages 318--328. Springer, 2024.

\bibitem[Xu et~al.(2024{\natexlab{a}})Xu, Usuyama, Bagga, Zhang, Rao, Naumann, Wong, Gero, González, Gu, Xu, Wei, Wang, Ma, Wei, Yang, Li, Gao, Rosemon, Bower, Lee, Weerasinghe, Wright, Robicsek, Piening, Bifulco, Wang, and Poon]{xu2024gigapath}
Hanwen Xu, Naoto Usuyama, Jaspreet Bagga, Sheng Zhang, Rajesh Rao, Tristan Naumann, Cliff Wong, Zelalem Gero, Javier González, Yu Gu, Yanbo Xu, Mu Wei, Wenhui Wang, Shuming Ma, Furu Wei, Jianwei Yang, Chunyuan Li, Jianfeng Gao, Jaylen Rosemon, Tucker Bower, Soohee Lee, Roshanthi Weerasinghe, Bill~J. Wright, Ari Robicsek, Brian Piening, Carlo Bifulco, Sheng Wang, and Hoifung Poon.
\newblock A whole-slide foundation model for digital pathology from real-world data.
\newblock \emph{Nature}, 2024{\natexlab{a}}.

\bibitem[Xu et~al.(2024{\natexlab{b}})Xu, Wang, Yang, Li, Ma, Chen, Wang, Huang, Gould, Lu, et~al.]{xu2024stomicsdb}
Zhicheng Xu, Weiwen Wang, Tao Yang, Ling Li, Xizheng Ma, Jing Chen, Jieyu Wang, Yan Huang, Joshua Gould, Huifang Lu, et~al.
\newblock Stomicsdb: a comprehensive database for spatial transcriptomics data sharing, analysis and visualization.
\newblock \emph{Nucleic acids research}, 52\penalty0 (D1):\penalty0 D1053--D1061, 2024{\natexlab{b}}.

\bibitem[Yang et~al.(2022)Yang, Wang, Wang, Fang, Tang, Huang, Lu, and Yao]{yang2022scbert}
Fan Yang, Wenchuan Wang, Fang Wang, Yuan Fang, Duyu Tang, Junzhou Huang, Hui Lu, and Jianhua Yao.
\newblock scbert as a large-scale pretrained deep language model for cell type annotation of single-cell rna-seq data.
\newblock \emph{Nature Machine Intelligence}, 4\penalty0 (10):\penalty0 852--866, 2022.

\bibitem[Yin et~al.(2024)Yin, Liu, Zhou, Wong, and Yuen]{yin2024prompting}
Chong Yin, Siqi Liu, Kaiyang Zhou, Vincent Wai-Sun Wong, and Pong~C Yuen.
\newblock Prompting vision foundation models for pathology image analysis.
\newblock In \emph{Proceedings of the IEEE/CVF Conference on Computer Vision and Pattern Recognition}, pages 11292--11301, 2024.

\bibitem[Yuan et~al.(2023)Yuan, Pan, Zhao, Zhao, Xu, Li, Zhao, Zhang, and Yao]{yuan2023sodb}
Zhiyuan Yuan, Wentao Pan, Xuan Zhao, Fangyuan Zhao, Zhimeng Xu, Xiu Li, Yi Zhao, Michael~Q Zhang, and Jianhua Yao.
\newblock Sodb facilitates comprehensive exploration of spatial omics data.
\newblock \emph{Nature Methods}, 20\penalty0 (3):\penalty0 387--399, 2023.

\bibitem[Zhang et~al.(2024{\natexlab{a}})Zhang, Schroeder, Yan, Yang, Hu, Lee, Cho, Susztak, Xu, Feldman, et~al.]{zhang2024inferring}
Daiwei Zhang, Amelia Schroeder, Hanying Yan, Haochen Yang, Jian Hu, Michelle~YY Lee, Kyung~S Cho, Katalin Susztak, George~X Xu, Michael~D Feldman, et~al.
\newblock Inferring super-resolution tissue architecture by integrating spatial transcriptomics with histology.
\newblock \emph{Nature biotechnology}, pages 1--6, 2024{\natexlab{a}}.

\bibitem[Zhang et~al.(2024{\natexlab{b}})Zhang, Xu, Chen, Xie, and Chen]{zhangprototypical}
Yilan Zhang, Yingxue Xu, Jianqi Chen, Fengying Xie, and Hao Chen.
\newblock Prototypical information bottlenecking and disentangling for multimodal cancer survival prediction.
\newblock In \emph{The Twelfth International Conference on Learning Representations}, 2024{\natexlab{b}}.

\bibitem[Zhang et~al.(2024{\natexlab{c}})Zhang, Zhao, Duan, Liu, Zheng, Liang, Zhang, and Li]{zhang2024pathology}
Zeyu Zhang, Yuanshen Zhao, Jingxian Duan, Yaou Liu, Hairong Zheng, Dong Liang, Zhenyu Zhang, and Zhi-Cheng Li.
\newblock Pathology-genomic fusion via biologically informed cross-modality graph learning for survival analysis.
\newblock \emph{arXiv preprint arXiv:2404.08023}, 2024{\natexlab{c}}.

\bibitem[Zhao et~al.(2023)Zhao, Zhang, and Nie]{zhao2023large}
Suyuan Zhao, Jiahuan Zhang, and Zaiqing Nie.
\newblock Large-scale cell representation learning via divide-and-conquer contrastive learning.
\newblock \emph{arXiv preprint arXiv:2306.04371}, 2023.

\bibitem[Zheng et~al.(2023)Zheng, Chen, Ding, Wong, and Cheung]{zheng2023aquila}
Yimin Zheng, Yitian Chen, Xianting Ding, Koon~Ho Wong, and Edwin Cheung.
\newblock Aquila: a spatial omics database and analysis platform.
\newblock \emph{Nucleic Acids Research}, 51\penalty0 (D1):\penalty0 D827--D834, 2023.

\end{thebibliography}
}
\end{document}